\documentclass{article}
\usepackage{graphicx} 
\usepackage{cite}
\usepackage{hyperref}
\title{Federated Weather Modeling on Sensor Data}
\author{Shengchao Chen, Guodong Long\\
Australian AI Institute, Faculty of Engineering and IT, \\ University of Technology Sydney, Australia}
\date{Email: shengchao.chen.uts@gmail.com; guodong.long@uts.edu.au}

\begin{document}

\maketitle

\section*{Definition}
\begin{figure}[tbh]
    \centering
    \includegraphics[width=1\textwidth]{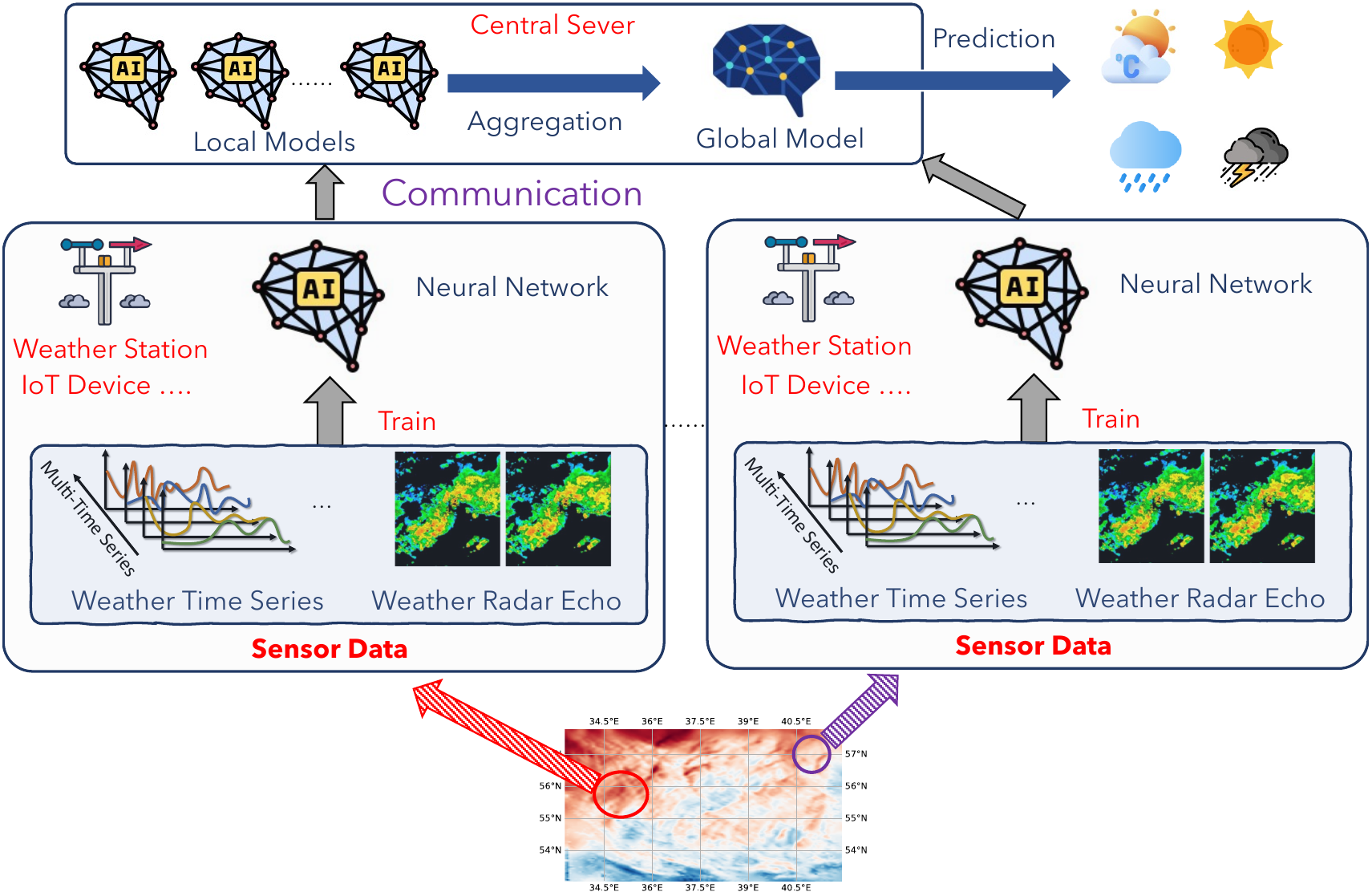}
    \caption{A simple schematic diagram of Federated Weather Modeling on Sensor Data (example of Ground Weather Station or IoT Device).}
    \label{fig:scheme}
\end{figure}

Federated weather modeling on sensor data (refer to \textbf{Figure~\ref{fig:scheme}}) is a distributed system underpinned by federated learning, enabling multiple sensor data sources, including ground weather stations, satellites and IoT devices, to collaboratively train deep learning models without sharing raw data. This method safeguards data privacy and security while leverages diverse, geographically distributed datasets to improve the accuracy and robustness of global/regional weather modeling tasks such as forecasting and anomaly detection.

\section*{Historical Background}
Climate change brings about significant, long-term shifts in global weather patterns. As surface temperatures rise this century, we can expect more frequent and severe extreme weather events. Advanced deep learning (DL) techniques applied to weather sensor data can improve our understanding of these changes and help develop targeted mitigation strategies~\cite{chen2023foundation}. Accurate predictions of sea-level rise are crucial for coastal urban planning and disaster preparedness, while short-term forecasts of rainfall and temperature are essential for public safety in sectors like agriculture and transportation.

Over the past decade, the use of data-driven DL in weather modeling has surged due to the increasing availability of global weather data~\cite{schultz2021can}. The primary challenge now is not just gathering data, but effectively integrating and utilizing various observational datasets to improve localized weather predictions. A common method involves combining data from multiple sources to train strong models. However, this approach encounters several challenges: (1) transmitting large-scale weather data in real time can overload low-resource devices; (2) data privacy concerns prevent the merging of sensitive information, leading to gaps in insights; and (3) geographic differences in data distribution can introduce biases in model outcomes.

Federated Learning (FL)~\cite{mcmahan2017communication} offers a promising alternative for weather modeling, addressing the limitations of centralized training. FL allows multiple data holders to collaborate on model training without sharing raw data, helping to overcome data silos and enabling precise, region-specific weather predictions on edge devices. This method is particularly valuable for regional weather modeling as it integrates insights from global data. Key focuses include understanding weather data through time series analysis for improved forecasting and developing FL methods to aggregate insights across regions to enhance regional models.

Early efforts in federated weather modeling aimed to improve global weather forecasting using standard time series models. For example, Farooq et al. ~\cite{farooq2023ffm} employed FL to predict floods by analyzing factors like snowmelt and rainfall runoff. Wen et al.~\cite{wen2022solar} developed a federated solar prediction model employing attention-based neural networks to enhance weather forecasting platforms, while Joel et al.~\cite{joel2024fedlstm} combined federated modeling with Radio Access Network (RAN) to improve weather data analytics across various stations. However, traditional time series models often struggle with the complex trends in weather data, and typical federated approaches frequently overlook the intricate spatial relationships between regions, which are vital due to the interconnected nature of atmospheric and oceanic conditions.

Recent research in federated weather modeling has shifted towards integrating spatial relationships to improve both global and regional forecasting. Chen et al.~\cite{chen2023prompt} applied graph-based techniques to better understand inter-device relationships by aggregating models from different weather devices. They also introduced a method that uses geographic location to tailor regional insights more effectively~\cite{chen2023federated}. Furthermore, advancements in large language models (LLMs) for time series analysis have led to a new approach in federated weather modeling, transitioning from traditional DL models to pretrained LLMs~\cite{chen2024personalized}. This approach leverages the similarities between language and weather modeling, both of which rely heavily on analyzing sequential data.

\section*{Scientific Fundamentals}
Preliminary for federated weather modeling include time series analysis, understanding weather data, and the application of federated learning specifically tailored to weather modeling.

\paragraph{Time Series Analysis} Time series is the most common format used in real-world applications, and it plays a vital role in weather modeling. This is because weather involves a sequence of data points across various meteorological factors~\cite{lim2021time}. Time series analysis aims to find specific insights by identifying patterns in these complex sequences, which can be applied to tasks like forecasting, classification, and detecting anomalies. There are two main types of techniques used in time series analysis: statistical methods and modern deep learning approaches. Traditional methods look at the inherent properties of the data, such as trends and seasonality, to understand patterns. In contrast, modern methods use deep neural networks, like Convolutional Neural Networks (CNNs)~\cite{wu2022timesnet}, Recurrent Neural Networks (RNNs)~\cite{lin2023segrnn}, and Transformers~\cite{wen2022transformers}, to automatically extract features from the data. While weather data is a type of time series, it has unique complexities due to its spatial and temporal patterns, as well as the influence of physical factors like interactions between variables and geographical features. As a result, general time series models often struggle with weather data and need to be adjusted to effectively handle these complexities.

\paragraph{Weather Data Understanding} Weather data understanding employs advanced machine learning techniques to derive decision-making insights from various sources, including spatio-temporal time series and relevant metadata, such as the geographic locations where weather data are collected~\cite{chen2023foundation}. This approach views weather modeling as a complex problem that involves time series analysis, spatio-temporal relationship modeling, and the integration of metadata. The goal is to create representations that can be used for different weather forecasting tasks. The main advantage of these techniques is their robustness, which surpasses traditional time series analysis methods by incorporating additional informative features like metadata. However, several challenges remain: (1) these techniques often overlook the distributed nature of data sources and the rapid growth in data volume, making traditional centralized training impractical due to high communication costs; (2) many weather devices are low-resource edge devices that cannot support the training of large-scale neural networks, which are essential for optimal performance; (3) data heterogeneity across different weather sources—caused by factors like geographic location and atmospheric conditions—can lead to decision-making biases in the trained models.

\paragraph{Federated Learning} Federated learning (FL)~\cite{mcmahan2017communication} is a distributed ML approach that enables collaborative model training across multiple participants while protecting data privacy. In FL, participants train models locally and share only model updates, not raw data. This method is particularly useful for handling sensitive information and allows learning from geographically diverse or privacy-sensitive sources. Additionally, FL accommodates data variability by allowing each client to create a customized model instead of relying on a single global model~\cite{tan2022fedproto,tan2022federated}. For example, Li et al.~\cite{li2021fedbn} introduced a personalized FL algorithm that keeps layers sensitive to data differences local for customization, while sharing other layers globally for knowledge exchange~\cite{arivazhagan2019federated}. In time series analysis, Chen et al.~\cite{chen2022personalized} used a Graph Neural Network to model spatial relationships and data distributions among participants, enhancing time series modeling with personalized features. Moreover, Li et al.~\cite{li2024fedasta} developed a framework that combines spatial dependencies and temporal relations using a masked spatial attention module and adaptive graph construction. Despite its advantages in distributed training and managing data variability, FL faces challenges: many models originally designed for image recognition or general time series data struggle with the unique characteristics of weather-specific data.

\paragraph{Federated Weather Modeling}
Federated weather modeling aims to build strong models by using insights from various regional weather data sources. This approach is promising for analyzing distributed and reliable weather data. As shown in Figure 1, it treats each weather station as an independent client that trains a model locally. This involves three key processes: (1) local client updates, (2) parameter communication, and (3) global aggregation. Among them, local client updates are the first step in the FL lifecycle. Clients start with an initial model from the server, refine it with local weather data, and then send the updated parameters back to the server. Recent studies recommend using time series Transformer architectures for local models to capture temporal trends in weather data. Moreover, parameter communication is crucial for aggregation, often involving clients sending updated parameters during each round. In cases of data variability, clients may share representations reflecting their local data distributions instead of full model parameters. Finally, global aggregation is essential for sharing knowledge among different sources. Traditional FL methods use weighted averaging, but this is less effective with significant data variability. Chen et al.~\cite{chen2023prompt} proposed a graph-based aggregation method to identify relationships among clients and improved the process by incorporating location data and physical constraints~\cite{chen2023federated}. Additionally, aggregating fewer parameters has emerged as an effective technique for efficient knowledge sharing.

\section*{Key Applications}

\paragraph{Specific Meteorological Variable Forecasting} Federated weather modeling can enhance weather forecasting by addressing specific meteorological variability based on different forecasting goals. By combining knowledge from various weather data sources, we can derive generalized insights that improve forecasts across different locations. For example, Vita et al.~\cite{de2024federated} integrated FL with clusters of automated weather stations to boost the accuracy of weather research and forecasting models. This integration aims to efficiently use diverse observational datasets for better local predictions. They treat each weather station as an independent computational node and apply deep time series analysis models for federated weather modeling. Additionally, Konda et al.~\cite{konda2024fedwavg} utilized FL with unmanned aerial vehicles (UAVs) to develop weather forecasting models based on cellular phone weather data. They introduced weighted federated averaging to address delays caused by outliers or inaccuracies, thereby increasing the reliability of forecasts under real-world conditions. The prediction of variables that assist weather forecasting, such as solar and wind data, has also been extensively studied using FL. For instance, federated weather modeling that incorporates radar imagery offers a new perspective for improving forecasts. Sainz et al.~\cite{sainz2024personalized} explored FL architectures with distributed weather radar imagery. They introduced a personalized FL strategy to tackle the challenges of rainfall forecasting with this data. 

\paragraph{Race or Extreme Events Detection} Detecting rare and extreme weather events is a major challenge in weather modeling, especially in federated weather modeling systems that deal with unbalanced data. Jafarigol et al.~\cite{jafarigol2023federated} used generative adversarial networks (GANs) within a FL framework to tackle the difficulties of forecasting in data-imbalanced situations, specifically focusing on rare event detection. They employed GANs for data augmentation to balance the data distribution across various independent sources and incorporated several oversampling techniques to improve the results. Kraatz et al.~\cite{kraatz2024federated} enhanced the disaster resilience of European communities by developing coordinated strategies for Disaster Risk Reduction and Climate Change Adaptation. Their project created an interoperable spatial information system that integrates multiple data sources and models to support flood risk modeling and address technical challenges like data interoperability. Sive et al.~\cite{siva2024federated} introduced a flood prediction model that uses FL to ensure data privacy. This model combines locally trained models from 18 different weather stations and takes regional factors into account to provide accurate five-day flood warnings.

\begin{figure}[tbh]
    \centering
    \includegraphics[width=1\textwidth]{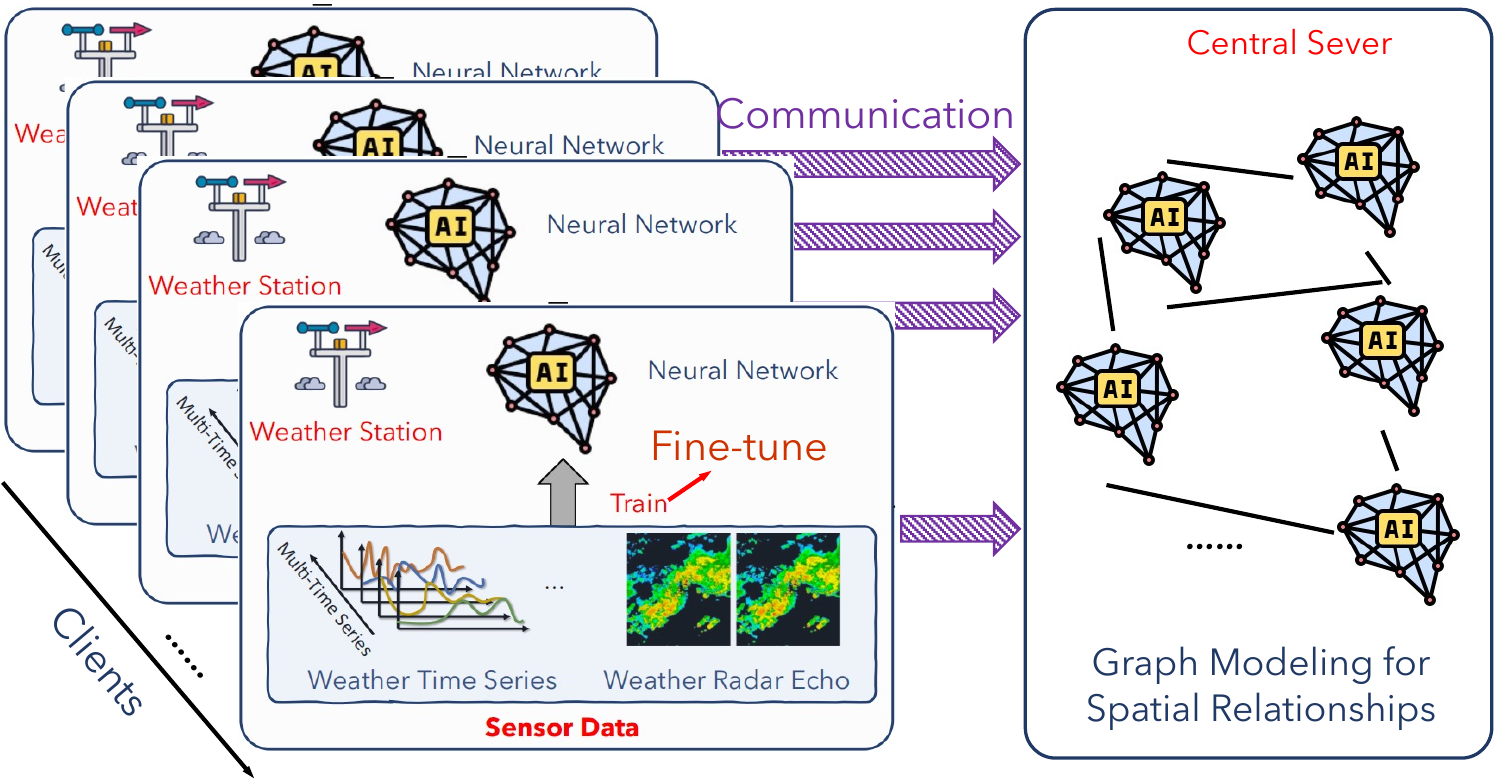}
    \caption{General-purpose Federated Weather Modeling.}
    \label{fig:general}
\end{figure}

\paragraph{General-purpose Weather Modeling} Imagine a unified weather model that can handle various tasks without the need for extensive design and training. This general-purpose weather model is a key goal in the field. However, creating such a model from scratch is challenging. It requires high-quality, large-scale weather data and a well-designed deep neural network. To overcome these challenges, researchers often use pretrained models, which allow them to build on existing knowledge and apply fine-tuning techniques for better results, like \textbf{Figure~\ref{fig:general}}. For instance, Chen et al.~\cite{chen2023prompt} suggested using partial data to pretrain a Transformer model, which they then fine-tuned in federated environments. This method demonstrated strong capabilities in forecasting multiple weather variables. Building on this, Chen et al.~\cite{chen2023federated} enhanced the model by adding location-related data and creating a graph to illustrate potential spatial relationships. However, these approaches typically depend on pretraining with available datasets, which can be a limitation if those datasets are insufficient. To tackle this problem, Chen et al.~\cite{chen2024personalized} introduced a method for fine-tuning pretrained large language models (LLMs) efficiently. This adapts LLMs for weather sequence modeling in federated environments, resulting in excellent performance for long-term forecasting, data imputation, and detecting anomalies.

\section*{Future Directions}
There are still many open challenges to be addressed in federated weather modeling. In particular, we summarize potential research directions as to improve the effectiveness and efficiency of federated weather modeling on sensor data. 

\paragraph{Interpretability} A major challenge in using deep learning models for federated weather modeling is understanding how these models make decisions. Many algorithms are complex and difficult to interpret, making it hard for users like researchers and weather forecasters to grasp their decision-making processes. While interpretability is less critical in areas like machine translation or text generation—where performance is often the main focus—it is vital in weather modeling~\cite{chen2023foundation}. Opaque models can lead to serious prediction errors, which may have harmful effects on society and the environment. For instance, it is important to know which meteorological factors influence forecasting decisions, how different weather variables interact, and why certain clients or data sources might be prioritized in the model's predictions. A reliable federated weather modeling system must provide clear explanations of its decision-making, ensuring transparency and building trust for real-world weather applications.

\paragraph{Multi-Modal Federated Weather Modeling across Domains} 
Weather time series data can be improved by adding extra information, such as text descriptions and images. This approach is especially useful in finance, where predictive models combine textual data from news articles and social media with economic time series. Similarly, weather forecasting benefits from integrating various data types, including reanalysis data and multimodal observational data like radar~\cite{sainz2024personalized} and satellite imagery~\cite{li2023fedfusion}. While it is relatively straightforward to implement this integration in centralized training environments, managing multimodal data in federated weather modeling is much more challenging. This difficulty arises from the differences in data types and their distributions across devices. Additionally, the performance of federated weather models can suffer when different types of data are unevenly distributed among devices.

\paragraph{Federated Weather Foundation Models} Foundation models (FMs), such as LLMs, have demonstrated impressive abilities to adapt across various tasks due to their extensive pre-training and large number of parameters~\cite{chen2023foundation,chenfedal,yan2025federated,chen2026bi,feng2025taming,feng2026visual}. While these models excel in fields like computer vision and natural language processing—where there is typically a lot of training data—weather modeling presents unique challenges. High-quality, large-scale weather data is not readily available, even though weather stations around the world continuously generate data. First, weather data often contains significant noise or interruptions, which can obscure important patterns. Second, many weather stations are located in sensitive areas, making it difficult to transmit data to centralized servers for training. Additionally, regional and topographical differences create significant variability in weather data, leading to potential biases during centralized model training. To address these issues, federated weather models offer a promising approach. They allow for decentralized training while maintaining data privacy~\cite{chen2024federated}. Furthermore, combining FL with mixture-of-experts (MoE) techniques could enhance the development of effective weather foundation models.

\bibliographystyle{plain}
\bibliography{ref}
\end{document}